\documentclass[11pt,a4paper]{article}
\usepackage[utf8]{inputenc}
\usepackage{booktabs}
\usepackage{caption}
\usepackage{cite}
\usepackage{fancyvrb}
\usepackage{graphicx}
\usepackage{hyperref}
\usepackage{makecell}
\usepackage[onehalfspacing]{setspace}
\usepackage{textcomp}

\newcounter{fragment}
\stepcounter{fragment}

\date{}

\begin{document}

\title{Human strategies for correcting `human-robot' errors during a laundry sorting task}
\author{
Pepita Barnard\,$^{1,*}$, Maria J Galvez Trigo\,$^{1,2}$, Dominic Price\,$^{1}$,\\
Sue Cobb\,$^{1}$, Gisela Reyes-Cruz\,$^{1}$, Gustavo Berumen\,$^{1}$,\\
David Branson III\,$^{1}$, Mojtaba A. Khanesar\,$^{1}$,
\\ Mercedes Torres Torres\,$^{4}$ and Michel Valstar\,$^{3}$\\ \\
$^{1}$University of Nottingham, Nottingham, UK \\
$^{2}$Cardiff University, Cardiff, UK\\
$^{3}$BlueSkeye AI, Nottingham, UK\\
$^{4}$B-Hive Innovations, Lincoln, UK}
\maketitle

\begin{abstract}
  Mental models and expectations underlying human-human interaction (HHI) inform human-robot interaction (HRI) with domestic robots. To ease collaborative home tasks by improving domestic robot speech and behaviours for human-robot communication, we designed a study to understand how people communicated when failure occurs.
  To identify patterns of natural communication, particularly in response to robotic failures, participants instructed Laundrobot to move laundry into baskets using natural language and gestures. Laundrobot either worked error-free, or in one of two error modes. Participants were not advised Laundrobot would be a human actor, nor given information about error modes.
  Video analysis from 42 participants found speech patterns, included laughter, verbal expressions, and filler words, such as ``oh'' and ``ok'', also, sequences of body movements, including touching one's own face, increased pointing with a static finger, and expressions of surprise. Common strategies deployed when errors occurred, included correcting and teaching, taking responsibility, and displays of frustration. The strength of reaction to errors diminished with exposure, possibly indicating acceptance or resignation.
  Some used strategies similar to those used to communicate with other technologies, such as smart assistants. An anthropomorphic robot may not be ideally suited to this kind of task. Laundrobot's appearance, morphology, voice, capabilities, and recovery strategies may have impacted how it was perceived. Some participants indicated Laundrobot's actual skills were not aligned with expectations; this made it difficult to know what to expect and how much Laundrobot understood. Expertise, personality, and cultural differences may affect responses, however these were not assessed.
  \tiny
\end{abstract}

\section{Introduction}
Challenges with home tasks can be compensated through technology, such as domestic robots \cite{Beer2012}, however robots require increasingly sophisticated interfaces to attend to human users' communication preferences. Since robots in Human Robot Interaction (HRI) are social agents, they elicit mental models and expectations known from human-human interaction (HHI) \cite{Mirnig2017}. Humans utilise multimodal social signals to interact and coordinate with each other smoothly, and without obvious effort; these signals include speech, body movements such as gestures, manipulations of objects, and combinations of these~\cite{Loth2016}. Social signal processing is the domain within computing that addresses social signals in both human-human and human-machine interactions~\cite{burgoon2017social}.
Identifying the content of social signals requires combining and interpreting information from several modalities~\cite{Loth2016}. Given the growing evidence of commomalities between human-human communication and human-robot communication, HHI research plays a key role as both a source of inspiration and a benchmark for embodiment of HRI~\cite{Esmer2022, Lagerstedt2020, Lamb2018}.

How people naturally communicate with robots in domestic settings is a growing area of interest, given that robots need to respond appropriately to humans to effectively engage in bi-directional communication~\cite{McColl2016}.
Previous studies underscore the importance of an in-depth understanding of human behaviours for successful long-term HRI in domestic settings~\cite{Weiss2017}.
Though some users tend to anthropomorphise robots, task-orientated and transactional conversations, such as those had with acquaintances or strangers in more limited service-oriented encounters, may be a good starting metaphor for structuring social agent conversation for HRI \cite{Clark2019, Bangalore2008}.
As they will inevitably fail~\cite{Knepper2017}, robots need capabilities related to recovery from errors and failure to help manage their interactions with people~\cite{Brooks2016}.

To inform the development of domestic robot speech and behaviours, and with the aim of improving human-robot communication in collaborative tasks, we designed a study to understand how people communicate with a `trainee robot' when robot failure occurs during a pick-and-place task with laundry items. Professional human actors were employed to emulate the `trainee robot' because they are skilled to consistently communicate characters and situations to their audience using controlled speech, body language and movement \cite{Bremers2023}. We will refer to `trainee robot' as Laundrobot, a name suggested by several participants.

The study required participants to instruct Laundrobot to move twelve items of clothing, one at a time, from a starting location into one of four baskets positioned around Laundrobot. The study comprised three conditions; one in which Laundrobot correctly followed instructions as given by the participants, and two in which it performed deliberate errors, with the participant being required to instruct it to make corrective actions. In the first error mode, Laundrobot did not cooperate in recovering misplaced items, instead it waited for participants to return items to the original location and repeat the instruction. In the second error mode, Laundrobot did cooperate by recovering items itself when it was made aware of the error, being compliant and helpful.

The purpose of the study was to capture and analyse participant's verbal and gestural instructions to identify patterns of natural communication, particularly in response to Laundrobot's failure to accurately follow the intended instruction, to inform the development of future social robot interaction during collaborative tasks.

\section{Related Work}
Social signals are \textit{observable} behaviours that people display during social interactions that have an effect on others; they are not random, rather they follow \textit{principles and laws}~\cite{burgoon2017social}.
Bremers and colleagues' 2023 exploration of current research addressing robots' use of social cues to recognise task failures in HRI~\cite{Bremers2023} found a variety of complex human reactions to human failures, including body movement, verbalisations, gaze, and facial expressions.
In 2015, video data of 137 participants from 5 HRI studies were systematically analysed for social signals produced by humans during problematic situations with robots~\cite{Giuliani2015}. From these video data analyses, technical failures, social norm violations, and verbal and non-verbal social signals were distinguished and categorised e.g., speech, gaze, head orientation, and body posture~\cite{Giuliani2015}. Pick-and-place tasks are central to the development of robust and adaptive HRI and HRC, partly due to their ubiquity in daily life~\cite{Lamb2019}. Behaviours and interaction patterns emerge from the dynamics of interacting systems composed of two humans or a human and a robot embedded in a task space~\cite{Lamb2018}.

To capture participants' response to physical pre-programmed unexpected robot errors while ``programming" a robot to unpack two boxes of groceries from a crate (a pick and place task), facial action units of 23 participants were logged~\cite{Stiber2022}; in a smaller related study, seven participants were studied under similar conditions with the addition of social signals~\cite{Stiber2022}.
Participants were blinded to the fact that their ``programming" had no effect on the robot's behaviour and errors in these Wizard of Oz (WoZ) studies~\cite{Stiber2022}, i.e. studies where the experimental researcher(s) acts as a ``Wizard", giving participants the impression they are interacting with a computer programme~\cite{kelley83}.

Mirnig and colleagues conducted a WoZ study with 45 participants, 21 of whom interacted with an erroneous robot during a conversation and a LEGO\textregistered~brick task, capturing social signals participants displayed towards the robot~\cite{Mirnig2017}.
Human-robot collaboration (HRC) studies are concerned with environments where humans and robots share tasks, occupied space, and resources with mutual awareness of each other's current and foreseen behaviour~\cite{Kardos2018}.
In a WoZ HRC study with 50 participants using a dual-arm cobot, humans' social signals were analysed during situations with purposefully executed erroneous robot behaviours and error-free situations~\cite{Cahya2019}.
Preliminary HHI investigations, with humans in place of robots, have helped inform protocol designs for HRI studies~\cite{Ivaldi2017}.

\section{Approach}
In this laboratory-based HHI study we sought to identify verbal and non-verbal communications elicited by humans in response to errors. Professional actors playing Laundrobot acted in one of three modes, each with rules related to errors and recovery actions. The objective was to identify speech and gestures used by the participants during the laundry sorting task, with particular interest in those occurring in response to Laundrobot's errors, recovery, and impacting task completion.

\subsection{Study design}
\begin{figure}[ht]
  \centering
  \includegraphics[width=\linewidth]{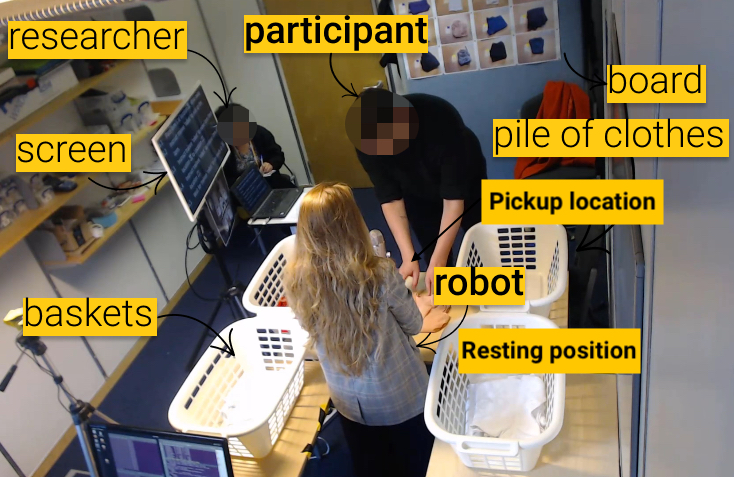}
  \caption{Study setup including symbols used to designate locations. Participant shown placing an item of clothing in the position designated pick up point for Laundrobot.}
  \label{fig:study}
\end{figure}

The study required participants to collaborate with Laundrobot to sort twelve items of clothing into four baskets.
Laundrobot stood in the centre of a set of tables and was surrounded by four baskets (Figure~\ref{fig:study}). The participant stood facing Laundrobot and picked each item, one piece at a time, from the pile of clothes to their left. One researcher was tasked with observing the participants and responding to any questions they had during the study.

In preparation for the study, the researchers provided the actors with detailed description of the study design, protocol procedure, their role, and behavioural affordances and constraints that they should emulate while in character.
The study procedure was piloted with the actors to confirm their understanding and performance requirements.
Laundrobot wore mirrored sunglasses to reduce the chances of eye contact with the participant, given eyes are more likely than the lower face to produce spontaneous displays as opposed to intentional expressions~\cite{burgoon2017social}.
Participants interacted with Laundrobot in one of three modes. Each mode reflected Laundrobot's response to failures:

\begin{itemize}
  \item \textit{Mode 1 - Gradual improvement throughout task}: Laundrobot did not make any deliberate errors. If a genuine error occurred, the participant was expected to fix it.
  \item \textit{Mode 2 - Errors made, no help from Laundrobot to recover}: Laundrobot committed deliberate errors on items: 5, 7, 8, 10, 11, and 12. Laundrobot expected the participant to fix the errors.
  \item \textit{Mode 3 - Errors made, Laundrobot helps in recovery}: Laundrobot committed deliberate error on items: 5, 7, 8, 10, 11, and 12. Laundrobot collaborated with the participant to fix the errors.
\end{itemize}

\subsection{Procedure}
In the pre-experiment briefing, participants were informed they would be working with a `trainee robot' learning to sort laundry ready for washing. They were asked to use natural language and gestures to communicate their instructions. To elicit natural responses from participants when trying to correct errors, they were not told that a human actor would play the part, nor were they informed of Laundrobot's mode and expected behaviour.

During the task, participants followed instructions displayed on a screen to their right. These instructions specified which basket each item should be placed in. To elicit natural language descriptors from the participants, the instructions included a diagram labelling the start position, robot position and position of each basket using symbols as shown in Figure~\ref{fig:study}. This provided a visual reference to aid the participant to understand the task instructions.

Table~\ref{table:goals} provides an overview of the task goals and sequence of actions for the items of clothing. The participant picked up an item from the top of the clothing pile and placed it directly in front of them on the table. They then instructed Laundrobot to pick up the item, rotate in the appropriate direction so that the item was above the correct basket and then drop the item in the basket.

\begin{table}[ht!]
  \centering
  \vspace{0.2cm}
  \caption{Goals/Sub-Goals and Actions}
  \label{table:goals}
  \footnotesize
  \begin{tabular}{@{}p{0.3\linewidth}p{0.3\linewidth}p{0.3\linewidth}@{}}
    \toprule
    GOAL & SUB-GOAL & ACTIONS (in order) \\ \midrule
    Move item from pickup location to target basket according to clothing type.&
    \newlength\breakheight
    \settoheight{\breakheight}{A\linebreak}
    \vspace{-\breakheight}
    \begin{itemize}
    \item Move Laundrobot from its resting position to pickup location.
    \item Pick up item at pickup location.
    \item Move item from pickup location to above the defined basket.
    \item Release item into basket.
    \item Return Laundrobot to resting position.
    \end{itemize} &
    \textbf{Participant:}
    \begin{itemize}
    \item Indicate what item to pick up and where to pick it from.
    \item Indicate where to move the item to.
  \end{itemize}
    \textbf{Learner:}
    \begin{itemize}
    \item Activate `pick-up' action.
    \item Activate `move to defined basket' action.
    \item Activate `release' action.
    \item Activate `return' action.
  \end{itemize} \\
    \bottomrule
  \end{tabular}
\end{table}

Sessions began when participants started reading the instructions and ended when Laundrobot placed the last item. Sessions were recorded using two video cameras positioned toward the participants from above and their right-hand side. Measures of affect, e.g. facial action units, were not collected.

After the experiment, participants provided feedback on their experience and completed a post-experimental questionnaire to evaluate the perception and interaction with Laundrobot, adapted from the `Post-experimental questionnaire for evaluation of the human-humanoid collaborative tasks with physical interaction'~\cite{Ivaldi2017}.  They were also asked to suggest a name for the robot.
A post-study interview was conducted with participants to gain feedback regarding their attitudes towards Laundrobot and demographic information. We did not capture ethnicity, nor other cultural background data.

The study was approved by the University of Nottingham's Computer Science Ethics Committee, and informed consent was obtained from all participants. Participants were offered a thank-you £10 gift card.

\subsection{Participants}
The sample consisted of 42 participants whose ages ranged from 18 to 51 with a mean of 29.6 years (\textit{SD} = 7.8). Twenty-two of these participants self-identified as female, nineteen as male and one as questioning. The participants were randomly assigned to one of the three different robot modes (with a roughly even distribution of participants between conditions): 16 participants to \textit{mode 1}, 13 to \textit{mode 2} and 13 \textit{mode 3}, respectively.

\subsection{Data collection and analysis}
Video analysis was conducted to identify instances of speech and behaviour actions for each participant, in each mode, and each clothing item sorted.

The categories for coding participants' embodied actions and social signals from the video recordings were formed by adapting those presented by Giuliani and colleagues~\cite{Giuliani2015} and Wobbrock and colleagues~\cite{Wobbrock2009}. Data was coded into categories related to \textit{speech}, \textit{head and facial expressions}, and \textit{body movements}, as shown in the supplementary materials.

Data analysis included the number of instances observed and the duration of each interaction, together with relevant comments from the researcher observation notes to provide further explanation of specific interactions.

\subsection{Error types}
The Laundrobot actors were given instructions, including details about activating deliberate failures and recovery strategies in response to participants' instructions for specific clothing items when in Mode 2 and Mode 3. They were free to commit a variety of errors as long as participants were not required to touch or take things from Laundrobot to resolve the problem.

The most common error types were:
\begin{itemize}
  \item \textit{Failing to drop item in correct basket:} The Laundrobot dropped an item of clothing in either the wrong basket, or in another place such as the edge of a basket.
  \item \textit{Picking up wrong item:} The Laundrobot picked up an unrequested item from the table or the baskets.
  \item \textit{Not understanding instructions:} The Laundrobot failed to understand the command of the participant and did something different from what it was asked to do.
\end{itemize}

The most common error was ``dropping an item in the wrong basket'', and most of the patterns presented in the \textit{Findings} section are based on this error type.

\section{Findings}

\subsection{General features}
In total 7729 speech instances were observed across the study with 75\% of these made by participants (5800), a small number made by the researcher observing the study, usually to answer a query from the participant (5.5\% = 423 instances), and the remainder made by Laundrobot in communication with the participant (19.5\% = 1506 instances). Our analysis focused on the language and accompanying gestures used by participants to communicate their instructions to Laundrobot, and their responses to, and attempts to recover from, errors made by Laundrobot.

Regarding \textit{speech type}, the majority were statements (41.3\% = 3199 instances), questions (26.4\% = 2046 instances), affirmations (12.9\% = 999 instances) and discourse markers, such as ``uh", ``ok" and ``mm" (8.4\% = 650 instances). In these categories, considerably more instances were observed in `error mode' conditions than in `error-free mode'.

Likewise, some physical actions were frequently displayed by participants, whereas others were observed on only a few instances. The number of instances of each action differed between modes, sometimes with considerably more instances in the `error modes' than in the `error-free mode', and vice versa.

Behavioural actions that occurred most often and in all study conditions were identified (see table "Instances of Participants' Behaviours" in the supplementary materials); deeper analysis of speech and behaviour responses per condition and per clothing item revealed unique patterns in response to Laundrobot failures.
For example, the response \textit{Laugh} occurred only 133 times across the study, but they were noted more often in response to Laundrobot errors. In the following subsections we present verbal and non-verbal communication behaviours observed in response to Laundrobot's errors.

\subsubsection{Pattern repetition}
\begin{figure}[ht]
  \centering
  \includegraphics[width=\linewidth]{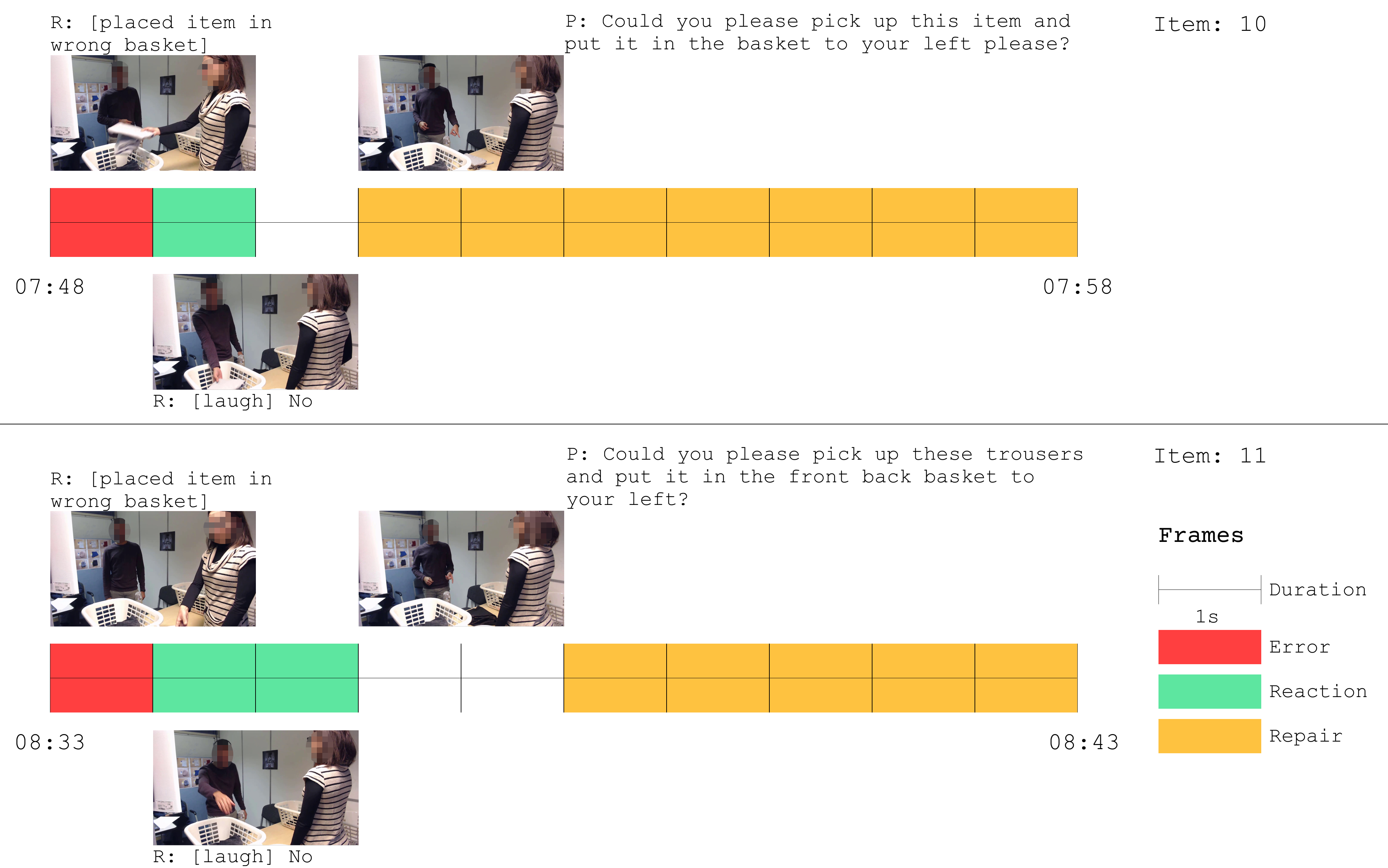}
  \caption{Example of a participant's~(P) pattern repetition in response to errors committed by Laundrobot~(R).}
  \label{fig:repetition}
\end{figure}

Some participants displayed a sequence of verbal commands and body movements that they repeated to some extent alongside the errors committed by Laundrobot. Examples of such patterns included the immediate use of specific filler words such as ``oh'', ``ok'', and ``mmm'', as well as sequences of body movements and verbal expressions such as touching one's own face and showing surprise. As an example, P116 (\textit{mode 2}), had a similar reaction to Laundrobot placing an item of clothing in the wrong basket in two distinct occasions (\textit{item 10} and \textit{item 11}); they laughed immediately after the error, said ``no", picked up the item from the wrong basket, then politely asked Laundrobot to put the item in the correct basket (Figure~\ref{fig:repetition}).

\subsubsection{Desensitisation to failure}
As participants were exposed to the errors committed by Laundrobot, the strength of their reaction diminished. Usually, the participants' reaction to the first error was greater than their reaction to the last error. Participants seemed to get used to error and became less surprised with each instance, showing some degree of habituation and desensitisation. An example of this was P139 (\textit{mode 2}). The reaction of the participant to the first error (\textit{item 5}) included laughing, bending towards Laundrobot, and smiling, while their reaction to the fourth error (\textit{item 10}) was only to smile and say, ``ok''.

\subsection{Strategies for correcting errors}
Participant reactions to the errors ranged from laughing loudly to having no reaction at all. Despite the diversity, there were some common strategies deployed, such as correcting and teaching, taking responsibility for the error, and indications of frustration.

\subsubsection{Correcting and teaching}
Participants took the mistakes committed by Laundrobot as an opportunity to instruct Laundrobot on how to do the task and improve its performance. Participants resorted to talking noticeably louder, repeating and clarifying instructions and even attempting to explain to Laundrobot the nature of the error. The latter was exemplified by P137 (\textit{mode 2}) after Laundrobot threw a piece of clothing (\textit{item 7}) to them instead of putting it in a basket:
\begin{Verbatim}[
  baselinestretch=1.5,
  vspace=0pt,
  fontsize=\footnotesize,
  =Text,
  label=Fragment \thefragment,
  commandchars=\\\{\}]
  \textit{(Error) Laundrobot threw item of clothing to participant}
  (1) P: (laugh) I\textquotesingle{}m not a basket, this is a basket
  (2) L:         You\textquotesingle{}re not a basket?
  (3) P: (laugh) Correct, this is a basket, and this is a basket
\end{Verbatim}
\stepcounter{fragment}

There were also instances when participants anticipated that an error was about to be committed and prevented Laundrobot from making the error. For example, P107 (\textit{mode 2}) said ``No, not that one" when they noticed that Laundrobot was about to drop (\textit{item 7}) in the wrong basket.

\subsubsection{Taking responsibility for the error}
Participants seemed to have attribution of the error displaying behaviours that suggest that they felt responsible for Laundrobot's unsuccessful actions. Participants often appeared nervous and laughed in disbelief as well as apologising and doubting themselves. The expression of apologetic words such as ``sorry", ``mind" and ``please" was more frequent in Mode 2 than Mode 3. Some participants doubted that the instructions they provided were accurate and that they had caused Laundrobot's deliberate error. For example, P137 (\textit{mode 3}) on (\textit{item 5}): ``I'm sorry (laugh) can you please take it out of that basket? (laugh)".

\begin{table}[h!]
  \centering
  \caption{Frequency of \textbf{apologetic} words in \textbf{error} and \textbf{no error} items.}
  \label{table:frequency}
  \footnotesize
  \begin{tabular}{@{}lll@{}}
    \toprule
    Words     & Error & No error \\ \midrule
    ``mind"   & 12    & 4        \\
    ``please" & 420   & 244      \\
    ``sorry"  & 31    & 15       \\ \bottomrule
  \end{tabular}
\end{table}

\subsubsection{Evidence of frustration}
Some participants did not display any immediate reaction to the errors committed by Laundrobot. We were not able to identify any difference in facial expressions or body movements for these participants in comparison to when Laundrobot was successful. These participants kept their composure and repeated the instructions in a clearer manner either by raising their voice or providing more detailed instruction. For example, P136 (\textit{mode 2}) for (\textit{item 5}) repeated the instructions after Laundrobot dropped the item in the wrong basket, providing a more detailed description of the target location ``in the basket on your right hand side that is nearest to you", including where not to put the item ``not the one far from you", and raising their voice in the second instance. Other participants converted to more precise movement instruction. For example, P109 (\textit{mode 3}) for (\textit{item 7}): ``Pick up the item again and put it in the one that's directly in front of that one 45 degrees".

Other examples show participants taking over from Laundrobot to do the task themselves, as seen by P126 (\textit{mode 2}) when Laundrobot failed to pick up (\textit{item 12}):
\begin{Verbatim}[
  frame=none,
  fontsize=\footnotesize,
  baselinestretch=1.5,
  vspace=0pt,
  =Text,
  label=Fragment \thefragment,
  commandchars=\\\{\}]
  (1) P: I've just put that in front of you, could you pick it up please?
  (2) P: Okay could you put this in the basket to your left which is \\
         closest to me here that I'm pointing at?
  \textit{(Error) Laundrobot failed to pick up item.}
  (3) P: Okay, okay. Could you - oh I tell you what I'll pick it up for you!
\end{Verbatim}
\stepcounter{fragment}

\subsection{Attitudes towards Laundrobot}
In general, participants had mixed feelings about the appearance of Laundrobot. Some indicated that because of this their expectations regarding its behaviour and capabilities were not aligned with its actual skills. Comments included that an anthropomorphic robot might not be the ideal type of robot for this kind of tasks, that it was difficult to know what to expect from Laundrobot and how much it understood, and that to communicate with it, they used strategies they would also use to communicate with other technologies, such as smart assistants.

\section{Discussion}
Despite diversity among participants' reactions to Laundrobot's errors, we report observed patterns in human speech and behaviour made by the participants, including laughter, use of verbal expressions and specific filler words such as “oh” and “ok” as well as sequences of body movements, including touching one's own face, more pointing with a static finger, and shows of surprise. Social signals have been shown to be effective and reliable sources of data for error detection, in spite of the greatly varied responses exhibited by people~\cite{Stiber2022}. Bremers and colleagues' literature review found that when people are able to see each other as failures occur they communicate emotions non-verbally, this includes a combination of gaze, facial expressions, nonverbal utterances, nodding, body position, and proxemics, amongst other gestures~\cite{Bremers2023}.

When in error situations with a robot, Guiliani and colleagues found participants smiled more, sometimes stopped moving, used many head movements, and in comparison to other social signals, used only a few hand gestures~\cite{Giuliani2015}; participants also looked back and forth between the robot and others in the room, or robot and objects,``literally looking for a solution to resolve the error situation"~\cite{Giuliani2015}.
Participants in Cahya and colleagues' study expressed more facial expressions, head gestures, and gaze shifts during erroneous situations than when in error-free situations~\cite{Cahya2019}. Facial expressions and gaze shifts had longer durations during error situations~\cite{Cahya2019}.
When Laundrobot was erroneous, our participants touched their own body, (e.g. face), stared, and pointed with a static finger pose more; they leaned towards a direction less than during non-error conditions.

In Mirnig and colleagues' WoZ study, most participants who experienced an erroneous robot exhibited clearly noticeable reactions to the robot's faults such as laughing, looking from task to the robot, annoyed facial expression~\cite{Mirnig2017}. Similarly, we identified more instances of laughter in error situations with Laundrobot. Additionally, we identified more instances of all speech types when Laundrobot was performing in the error modes.

No two participants reacted in the same way to the same error while ``programming" the grocery-unpacking robot~\cite{Stiber2022}; Stiber found the most common behaviours were smiling, talking, and looking away from the robot; rather than one set of expressions, reactions to errors involved a sequential evolution of facial movement. These behaviours were a series and escalated as time passed after the error~\cite{Stiber2022}.

Contrary to Stiber's findings, not all of our participants' reactions to errors escalated after an error. We noticed patterns in participants' responses to Laundrobot's repeated errors \cite{Stiber2022}. Some participants attempted to help Laundrobot learn when it made errors, some appeared to become desensitised to the repeated errors, displaying a diminishing reaction, possibly an indication of acceptance or resignation to the erroneous Laundrobot, whereas others displayed frustration by raising their voice in their repetitions.
Instances where our participants anticipated and stopped Laundrobot from committing an error were similar to how some of Stiber's participants responded to predictable robot errors.

Participants deployed some common strategies when errors occurred, including correcting and teaching, taking responsibility for the error, displays of frustration such as raising their voice, repetition, and provision of increasingly detailed descriptions and directions to Laundrobot.
Expertise and personality have indicated potential reasons for different responses to robot errors~\cite{Stiber2022}, however our participants' expertise and personality were not assessed. These aspects and questions addressing cultural differences may offer further insight into participants' differing responses.

A WoZ HRI study to complement our HHI study reported in this paper would explore how people respond to the Laundrobot when in the form of a collaborative robot arm. Such a WoZ study would be an opportunity to see if dynamics of social signals differ~\cite{Ivaldi2017}.

The range of effects that faulty robot behaviours cause is not yet full understood or agreed~\cite{Correia2018}. Previous research on human perception of a robot after an error situation has generated conflicting findings, suggesting that type of task, type of error, or severity of the error may be influencing factors~\cite{Correia2018}.
Other factors impacting how human's perceive a faulty robot may include the robot's capabilities, recovery strategies, appearance, morphology, and voice; additionally the domain and robot operating environment. Recovery from failure requires further exploration into how robot morphology, voice, and other characteristics may affect how humans experience HRI and HRC.

\section{Conclusion}
This HHI study sought to identify natural verbal and non-verbal communications elicited by humans in response to errors. Using a human actor to emulate a robot, we captured participants' speech and gesture patterns under different modes of robot compliance. Our study participants demonstrated a range of social signals similar to those found in HRI studies, including laughing, and smiling. Strategies for correcting errors included teaching, taking responsibility for errors, and displaying signs of frustration.

We therefore consider this method acceptable for identifying patterns in participants' responses to, and corrections of, robot errors.
These patterns can be used as indicators for a cobot to recognise when their human co-worker has noticed an error, and switch to a `learning mode'.
We plan to apply patterns identified here to a `library' of robot actions and responses for use in a replication study using a WoZ cobot. Further, these patterns of behaviour may be applied to autonomous cobots to provide effective methods of self-correcting erroneous behaviour when interacting with humans.

\section*{Funding}
This submission was supported by the EPSRC Grant Number EP/T022493/1 and EP/M02315X/1. Any enquiries regarding this submission should be sent to horizon@nottingham.ac.uk.

Released under the Creative Commons license: Attribution 4.0 International (CC BY 4.0): https://creativecommons.org/licenses/by/4.0/.

\section*{Supplementary Materials}\label{section:supplementary}

\newpage
\subsection*{Coding used in the Analysis of Video Data}
\begin{table}[ht!]
  \centering
  \vspace{0.2cm}
  \tiny
  \begin{tabular}{@{}p{0.3\linewidth}p{0.3\linewidth}p{0.3\linewidth}@{}}
    \toprule
    \textbf{Speech} &
    \textbf{Speech Producer}
    \begin{itemize}
      \item Participant
      \item Robot
      \item Researcher
    \end{itemize} &
    \textbf{Speech Type}
    \begin{itemize}
      \item Statement
      \item Address
      \item Question
      \item Laugh
      \item Negation
      \item Affirmation
      \item Correction
      \item Discourse Maker
      \item Other
    \end{itemize} \\ \midrule
    \textbf{Head and eye movements and facial expressions} &
    \textbf{Head/Eye Movement}
    \begin{itemize}
      \item\raggedright Turn to look at Laundrobot
      \item Turn to look at researcher
      \item Look at instructions
      \item Look at object
      \item Look in a direction
      \item Tilt head
      \item Nod
      \item Shake head
      \item Eye-gaze
      \item Stare
      \item Other
    \end{itemize} &
    \textbf{Facial expressions}
    \begin{itemize}
      \item Smile
      \item Grimace
      \item Raise eyebrows
      \item Frown
      \item Eyes wide-open
      \item Eyes half-closed
    \end{itemize} \\ \midrule
    \raggedright\textbf{Body posture and hand gestures} &
    \textbf{Body Posture}
    \begin{itemize}
      \item Lean towards Laundrobot
      \item Move towards Laundrobot
      \item Lean towards direction
      \item Move towards direction
      \item\raggedright Move away from Laundrobot
      \item\raggedright Turn or twist to align with Laundrobot
      \item Change body posture
      \item\raggedright Celebratory body movement
      \item Other body movement
    \end{itemize}&
    \textbf{Hand Gesture}
    \begin{itemize}
      \item Pointing with static finger pose
      \item Pointing with path
      \item Static hand pose
      \item Static hand pose following path
      \item Dynamic hand
      \item Dynamic hand pose with path
      \item Touch own body
      \item Emblem
    \end{itemize} \\ \bottomrule
  \end{tabular}
\end{table}

\newpage

\subsection*{Instances of Participants' Behaviours}\label{section:instances}
\begin{table}[ht!]
  \centering
  \tiny
  \begin{tabular}{@{}p{0.28\linewidth}p{0.1\linewidth}
      p{0.1\linewidth}p{0.1\linewidth}p{0.22\linewidth}@{}}
    \toprule
     & Mode 1\newline(n=16) & Mode 2\newline(n=13) & Mode 3\newline(n=13) & Total \newline(\% of category) \\ \midrule
     \multicolumn{5}{@{}p{\linewidth}@{}}{\textbf{Head/Eye Movement}} \\
     Stare & 330 & 443 & 490 & 1263 (17.2\%) \\
     Look in a direction & 540 & 364 & 533 & 1437 (19.6\%) \\
     Look at object & 818 & 1163 & 812 & 2793 (38.0\%) \\
     \raggedright Look at instructions & 740 & 596 & 514 & 1850 (25.2\%) \\ \midrule
     \multicolumn{5}{@{}p{\linewidth}@{}}{\textbf{Body Posture}} \\
     \raggedright Move away from Laundrobot & 123 & 175 & 105 & 403 (12.8\%) \\
     Move towards direction & 216 & 221 & 170 & 607 (19.4\%) \\
     Lean towards direction & 310 & 254 & 227 & 791 (25.3\%) \\
     \raggedright Move towards Laundrobot & 117 & 170 & 92 & 379 (12.1\%) \\
     Lean towards Laundrobot & 130 & 153 & 112 & 395 (12.6\%) \\ \midrule
     \multicolumn{5}{@{}p{\linewidth}@{}}{\textbf{Hand Gesture}} \\
     Touch own body & 113 & 164 & 200 & 477 (13.1\%) \\
     Static hand pose & 643 & 796 & 535 & 1974 (54.3\%) \\
     Pointing with static finger pose & 156 & 255 & 179 & 590 (16.2\%) \\
     \\ \bottomrule
  \end{tabular}
\end{table}

\bibliographystyle{acm}
\bibliography{strategies}

\end{document}